\documentclass[sigconf]{acmart}

\AtBeginDocument{%
  \providecommand\BibTeX{{%
    \normalfont B\kern-0.5em{\scshape i\kern-0.25em b}\kern-0.8em\TeX}}}

\copyrightyear{2026}
\acmYear{2026}
\setcopyright{acmcopyright}
\acmConference[MM '26]{Proceedings of the 34th ACM International Conference on Multimedia}{October 2026}{TBD}
\acmBooktitle{Proceedings of the 34th ACM International Conference on Multimedia (MM '26), October 2026, TBD}
\acmPrice{15.00}
\acmDOI{XXXXXXX.XXXXXXX}
\acmISBN{978-1-4503-XXXX-X/26/10}

\acmSubmissionID{3568}

\usepackage{booktabs}
\usepackage{amsmath}

\usepackage{amssymb}
\usepackage{graphicx}
\usepackage{multirow}
\usepackage{xcolor}
\usepackage{float}
\usepackage{tikz}
\usetikzlibrary{arrows.meta,positioning,decorations.pathreplacing}
\usepackage{amsthm}
\newtheorem{theorem}{Theorem}

\newtheorem{definition}[theorem]{Definition}

\usepackage{enumitem}
\usepackage{tcolorbox}
\usepackage{fvextra}
\usepackage{xcolor}
\tcbuselibrary{breakable, skins}

\newcommand{\ours}{\textsc{UIPress}}

\begin{document}

\title{\ours: Bringing Optical Token Compression \\ to UI-to-Code Generation}

\author{Dasen Dai}
\authornote{Equal contribution. Surname alphabetical order.}
\affiliation{\institution{The Chinese University of Hong Kong}\country{China}} 

\author{Shuoqi Li}
\authornotemark[1]
\affiliation{\institution{The Chinese University of Hong Kong}\country{China}} 

\author{Ronghao Chen}
\affiliation{\institution{Peking University}\country{China}} 

\author{Huacan Wang} 
\affiliation{\institution{University of Chinese Academy of Sciences}\country{China}}

\author{Biao Wu}
\affiliation{\institution{University of Technology Sydney}\country{Australia}}

\author{Qizhen Lan}
\affiliation{\institution{UTHealth}\country{USA}}


\begin{abstract}
UI-to-Code generation requires vision--language models (VLMs) to produce thousands of tokens of structured HTML/CSS from a single screenshot, making visual token efficiency critical.
Existing compression methods either select tokens at inference time using task-agnostic heuristics, or zero out low-attention features without actually shortening the sequence---neither truly reduces prefill latency or adapts to the non-uniform information density of UI screenshots.
Meanwhile, optical (encoder-side learned) compression has shown strong results for document OCR, yet no prior work has adapted this paradigm to UI-to-Code generation.
We propose \ours{}, a lightweight learned compression module inserted between the frozen ViT encoder and the LLM decoder of Qwen3-VL-8B.
\ours{} combines depthwise-separable convolutions, element-guided spatial reweighting, and Transformer refinement to compress ${\sim}$6{,}700 visual tokens to a fixed budget of 256.
Together with Low-Rank Adaptation (LoRA) on the decoder to bridge the representation gap, the entire system adds only ${\sim}$21.7M trainable parameters (0.26\% of the 8B base model).
Under a fair comparison on the same base model against four baselines on Design2Code, \ours{} at 256 tokens achieves a CLIP score of 0.8127, outperforming the uncompressed baseline by +7.5\% and the strongest inference-time method by +4.6\%, while delivering 9.1$\times$ time-to-first-token speedup.
To the best of our knowledge, \ours{} is the first encoder-side learned compression method for the UI-to-Code task.
\end{abstract}



\maketitle


\section{Introduction}
\label{sec:intro}

Automatically converting UI screenshots into functional HTML/CSS code (UI-to-Code, or UI2CODE) is a practical and increasingly important task in front-end engineering~\citep{si2024design2code, wu2024webcode2m}.
Unlike short-answer visual question answering (VQA)~\citep{deepseekai2025deepseekr1incentivizingreasoningcapability, llavaonevision,li2024llava,qwen2.5vl,gemini25pro,wang2025infinity,wu2026vision,fang2024large, shi2024human, shi2023self, shi2022stay, shi2025monte, shi2023cooperation}, UI2CODE requires generating 1{,}000--4{,}000 tokens of structured code that faithfully reproduces the spatial layout, text content, colors, and structural hierarchy of the input screenshot---making it one of the most demanding benchmarks for vision--language models (VLMs).
Modern VLMs represent screenshots as long visual token sequences~\citep{guiact,guig2,guiodyssey,GUI_Reflection,sun2025gui,chen2024guicourse,shi2025presentagent,wu2024foundations,yan2025automotive}. For example, Qwen3-VL-8B~\citep{qwen3vl2025} produces ${\sim}$6{,}700 visual tokens at native resolution on a typical Design2Code page, which dominate both prefill latency and GPU memory during autoregressive decoding. Reducing this visual token overhead is therefore critical for practical deployment.

A growing body of work addresses visual token reduction for VLMs~\citep{shang2024llavaprumerge,yao2024minicpmvgpt4vlevelmllm,tang2025magicguifoundationalmobilegui,wang2024eantlargescaledatasetefficient,li2024effectsdatascaleui,liu2023visualinstructiontuning,li2024llavaonevisioneasyvisualtask,yang2024visionzip,zheng2024gpt4visiongeneralistwebagent,dai2026papervoyager}.
Inference-time methods compress tokens without retraining: FastV~\citep{chen2024fastv} zeros out low-attention features after early layers; VisionZip~\citep{yang2024visionzip} selects dominant tokens by L2 norm and merges the rest; LLaVA-PruMerge~\citep{shang2024llavaprumerge} uses CLS attention for pruning.
These methods achieve impressive results on VQA benchmarks---FlashVLM~\citep{sun2025flashvlm} reports ``beyond-lossless'' performance at 78\% pruning on TextVQA and MMBench~\citep{liu2024mmbench}.
For UI2CODE specifically, EfficientUICoder~\citep{yun2024efficientuicoder} combines UI element detection with token selection, achieving 55--60\% compression on LLaVA-v1.6~\citep{li2024llavaonevisioneasyvisualtask,llavaonevision,xu2025llavacotletvisionlanguage,liu2023visualinstructiontuning}. In a separate line of work, DeepSeek-OCR~\citep{liu2025deepseekocr} introduces optical compression: a learned SAM+CLIP encoder that compresses document images into 73--850 tokens at the encoding stage, achieving 97\% OCR precision.

\begin{figure}[t!]
    \centering
    \includegraphics[width=\linewidth]{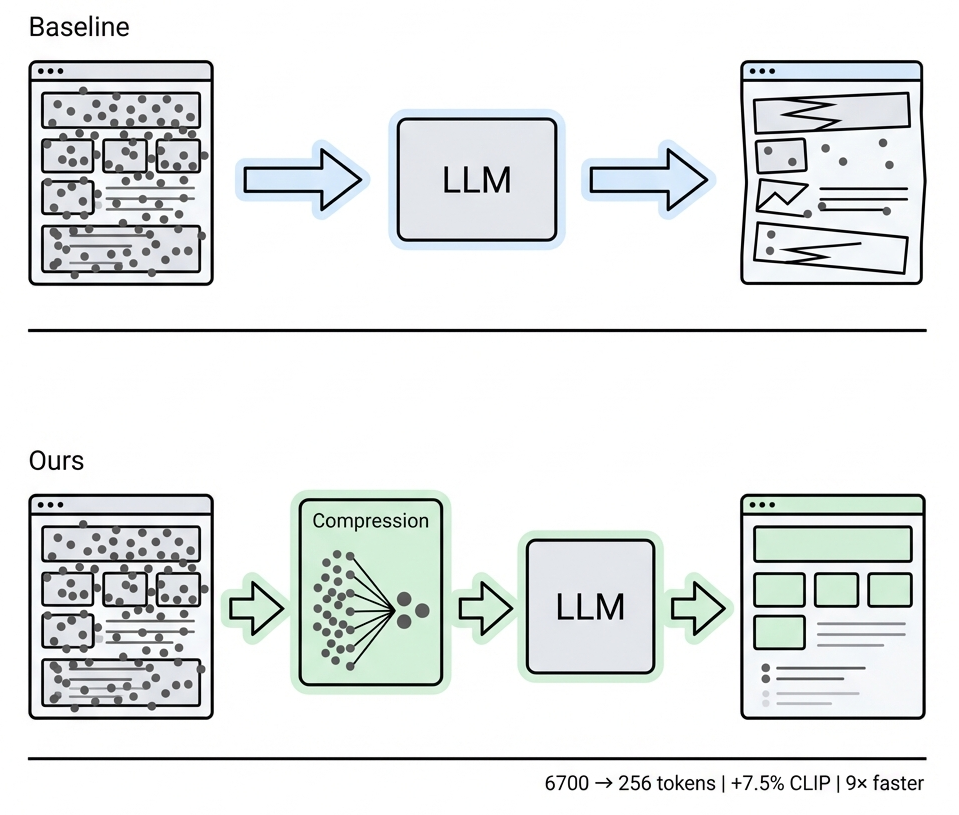}
    \vspace{-7mm}
    \caption{
    Comparison between conventional visual token processing and our learned compression framework for UI-to-Code generation. 
    }
    \label{fig:teaser}
    \vspace{-3mm}
\end{figure}

Despite this progress, existing methods face three fundamental limitations when applied to UI2CODE:
(i)~Feature-zeroing methods do not reduce computation. Approaches like FastV set pruned token features to zero while retaining their sequence positions. The full-length sequence still passes through every attention layer, yielding no reduction in time-to-first-token (TTFT) or peak VRAM---we verify this empirically in Section~\ref{sec:ttft}. (ii)~Selection heuristics are task-agnostic. VisionZip selects tokens by generic L2 norms and LLaVA-PruMerge uses CLS attention, neither of which accounts for the highly non-uniform information density of UI screenshots, where text regions and interactive elements carry far more semantic weight than background areas. (iii)~Optical compression has not been adapted to UI2CODE. DeepSeek-OCR's learned encoder achieves strong results on documents, but it uses a separate 1.3B-parameter encoder trained from scratch. Directly transferring its representations to a different VLM decoder fails due to embedding-space mismatch, and the OCR domain differs substantially from webpage rendering.

To address these challenges, we propose \ours{} (\textbf{UI} \textbf{P}ipeline for \textbf{R}epresentation-\textbf{E}fficient \textbf{S}creenshot \textbf{S}ynthesis), which brings optical compression to UI2CODE by training a lightweight compressor \emph{within} the target VLM's own representation space.
The key insight is that, rather than using a separate encoder or relying on post-hoc heuristics, we can learn a task-aware spatial compression directly on the visual tokens produced by the frozen ViT encoder of Qwen3-VL-8B. Concretely, \ours{} inserts an \emph{Optical Compressor} between the ViT encoder and the LLM decoder (Figure~\ref{fig:architecture}). The compressor applies depthwise-separable convolutions for spatial downsampling, element-guided reweighting (via OmniParser~\citep{lu2024omniparser} detections) to concentrate capacity on information-dense UI regions, and a Transformer refinement layer to recover inter-token dependencies---compressing ${\sim}$6{,}700 tokens to a fixed budget of $K{=}256$. To bridge the representation gap between compressed tokens and the frozen decoder, we apply Low-Rank Adaptation (LoRA)~\citep{hu2022lora} to the query and value projections of all decoder layers, adding only 7.7M trainable parameters. The compressor (14M parameters) and LoRA are trained jointly on 50K WebSight~\citep{huggingface2024websight} screenshot--HTML pairs via the standard autoregressive loss.

We evaluate \ours{} on Design2Code~\citep{si2024design2code} alongside four baselines under a unified protocol on the same base model (Qwen3-VL-8B).
At 256 tokens (25.5$\times$ compression), \ours{} achieves CLIP 0.8127, outperforming the uncompressed baseline by \textbf{+7.5\%}, the strongest inference-time method (resolution scaling, 845 tokens) by \textbf{+4.6\%}, and VisionZip at matched token count by \textbf{+10.8\%}, while delivering \textbf{9.1$\times$} TTFT speedup.
Ablations confirm that both the learned compressor and LoRA adaptation are essential, and the improvement holds across all page types.

To the best of our knowledge, \ours{} is the first encoder-side learned compression method for UI-to-Code generation.
In summary, our contributions are three-fold:
\begin{itemize}
    \item We propose \ours{}, a lightweight optical compression framework (${\sim}$14M parameters) that reduces visual tokens from ${\sim}$6{,}700 to 256 while \emph{improving} CLIP score by 7.5\% over the uncompressed baseline---demonstrating that learned compression can be ``beyond lossless'' for UI2CODE.
    \item We conduct a fair comparison on a single base model (Qwen3-VL-8B) across four compression paradigms: resolution scaling, token selection (VisionZip), element-guided pruning (EfficientUICoder), and feature zeroing (FastV), providing the first unified benchmark for visual token compression in UI2CODE.
    \item We provide comprehensive analysis including bootstrap confidence intervals, TTFT measurement, component ablations, and per-page-type breakdowns, offering actionable insights for future work on efficient UI understanding.
\end{itemize}

\section{Related Work}
\label{sec:related}

\subsection{UI-to-Code Generation and Visual Token Representations} 

Modern vision language models typically rely on large-scale pretrained visual encoders and language decoders that represent images as sequences of visual tokens corresponding to local patches~\citep{chu2024mobilevlm, radford2021clip,wu2024mmclip,mokady2021clipcap,vlmr1,wang2023cogvlm,vteam2025glm45vglm41vthinkingversatilemultimodal,visionr1,visual-rft,Vaswani+2017,yao2024minicpmvgpt4vlevelmllm}. Many systems adopt a modular architecture coupling a frozen visual encoder with a language model through a lightweight projection layer~\citep{mokady2021clipcap,li2023blip,llavaonevision,li2024llava,zheng2024llamafactory,grattafiori2024llama,qwen2.5vl,qwen3vl2025,guipivot}. Within this paradigm, Design2Code~\citep{si2024design2code} established the screenshot-to-code task using CLIP-based features with block-match and structural similarity metrics. Subsequent work like WebCode2M~\citep{wu2024webcode2m} has scaled training data to 2M samples to improve generalization. However, high-resolution images generate thousands of visual tokens, creating substantial overhead for cross-modal attention during decoding. EfficientUICoder~\citep{yun2024efficientuicoder} is the only prior work targeting visual token compression specifically for UI2CODE which combines UI element detection, attention refinement, and output-side deduplication on LLaVA-v1.6. While it achieves 55--60\% compression, much of the gain stems from output-side modules rather than input compression alone, leaving the efficiency of the visual representation unoptimized.

\subsection{Inference-time Visual Token Reduction}

A parallel research direction improves transformer efficiency by compressing intermediate representations during inference, particularly tokens or key--value caches. One common strategy identifies and retains influential tokens only~\citep{li2023deep,liu2025deepseekocr,yun2024efficientuicoder,li2024tokenpacker}. FastV~\citep{chen2024fastv} prunes tokens by attention score after early transformer layers, while VisionZip~\citep{yang2024visionzip} selects ``dominant'' tokens via encoder self-attention and merges the rest via clustering. Similarly, LLaVA-PruMerge~\citep{shang2024llavaprumerge} uses CLS attention for pruning followed by KNN merging, and FlashVLM~\citep{sun2025flashvlm} introduces text-guided selection with diversity-preserving partitioning. Other methods like Scissorhands observe that token importance persists across decoding steps and remove less significant entries from the cache accordingly~\citep{liu2023scissorhands}. Despite their effectiveness, these approaches operate primarily as inference-time optimizations applied after the encoder has produced dense token sequences, leaving the upstream visual representation unchanged~\citep{chen2024fastv}. Crucially, methods based on \emph{feature zeroing}, which sets pruned token features to zero while retaining their sequence positions, could not reduce prefill FLOPS or time-to-first-token (TTFT), as the full sequence still passes through the transformer~\citep{chen2024fastv}. Token reduction in these works is typically determined by architectural heuristics rather than an explicit optimization objective, resulting in the implicit compression fidelity trade-off.

\subsection{Optical Encoder-side Compression}

To address the limitations of post-hoc pruning, recent work explores compressing visual features in the projection stage before language decoding. TokenPacker aggregates global and regional information through a coarse-to-fine projection strategy~\citep{li2024tokenpacker}. In the domain of document understanding, DeepSeek-OCR~\citep{liu2025deepseekocr} trains a pipeline that compresses document images to 73--850 tokens, achieving strong OCR performance, while TextHawk2~\citep{yu2024texthawk2} co-trains the encoder and compressor for significant token reduction on OCR benchmarks. However, these works focus primarily on document or OCR tasks. We are the first to adapt the optical compression paradigm to UI-to-Code generation. In contrast to relying on fixed projection designs or post-hoc pruning, our approach frames token reduction as a learnable encoder-side compression problem. We learn a task-aware transformation that converts dense visual tokens into a compact representation optimized jointly with UI2CODE generation, enabling efficient multimodal processing while preserving the structural information required for accurate interface reconstruction.

\section{Methodology}
\label{sec:method}

\begin{figure*}[!htbp]
\centering
\includegraphics[width=0.95\textwidth]{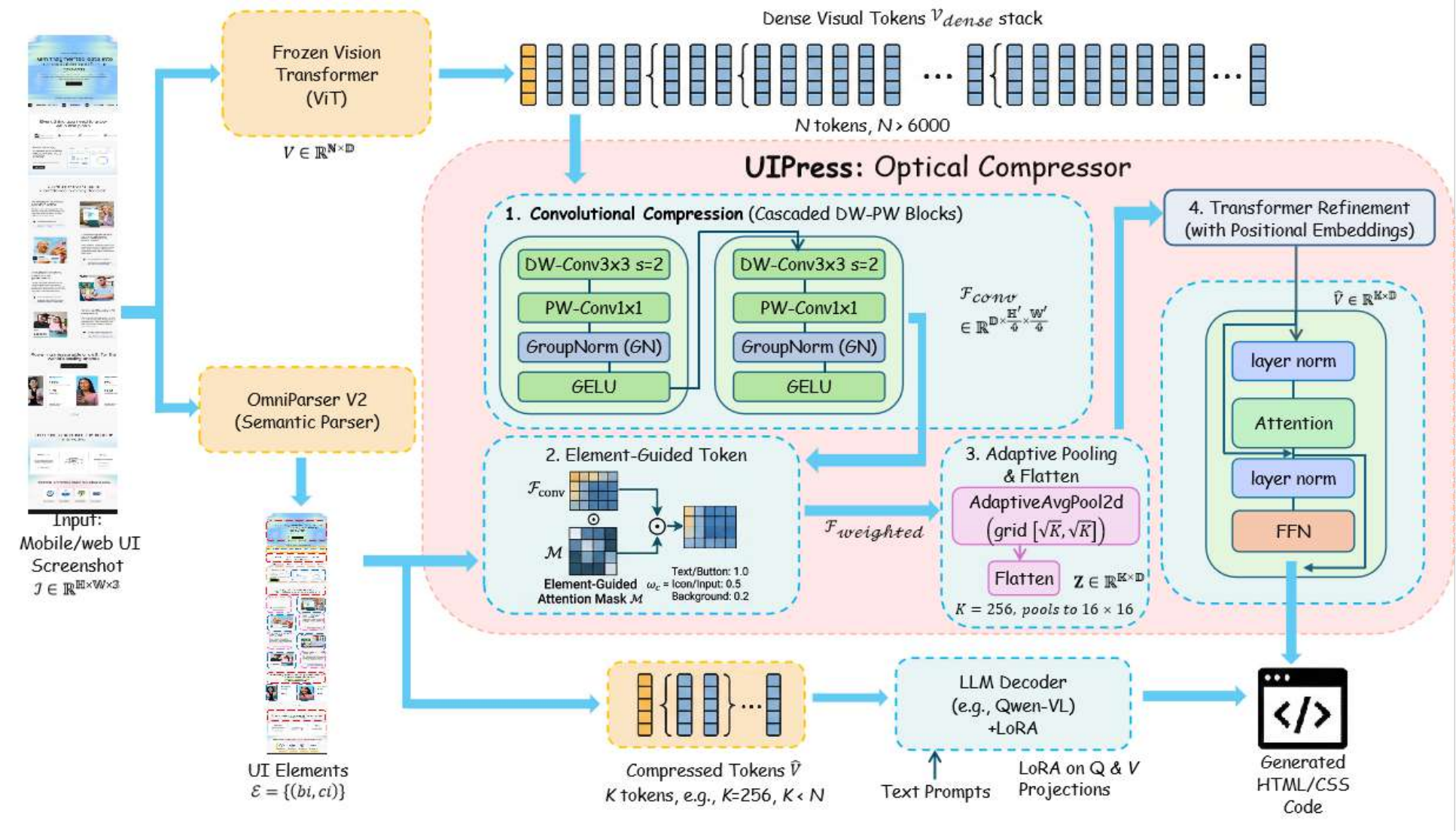}
\vspace{-5mm}
\caption{Overview of the \ours{} framework. A frozen ViT encoder produces $N{>}6{,}000$ dense visual tokens $\mathbf{V}$. The Optical Compressor consists of four stages: (1)~\emph{Convolutional Compression}---two cascaded depthwise-separable (DW) and pointwise (PW) convolution blocks with GroupNorm and GELU, yielding $4\times$ spatial downsampling; (2)~\emph{Element-Guided Token Reweighting}---an attention mask $\mathcal{M}$ derived from OmniParser~V2 assigns category-specific weights ($\omega_c{=}1.0$ for text/buttons, $0.5$ for icons, $0.2$ for background); (3)~\emph{Adaptive Pooling \& Flatten}---AdaptiveAvgPool2d reduces the weighted feature map to a $\sqrt{K}{\times}\sqrt{K}$ grid ($K{=}256$), then flattens to $\mathbf{Z} \in \mathbb{R}^{K \times D}$; (4)~\emph{Transformer Refinement}---a single-layer Transformer with learned positional embeddings recovers inter-token dependencies. The compressed tokens $\hat{\mathbf{V}}$ are decoded by the LLM augmented with LoRA adapters on Q and V projections.}
\label{fig:architecture}
\end{figure*}


As discussed in Section~\ref{sec:intro}, existing visual token compression methods either operate at inference time without reducing the actual sequence length, or rely on generic heuristics that ignore UI-specific structure.
In this section we present \ours{}, which addresses both limitations through a three-stage \emph{learned} compression pipeline inserted between the frozen visual encoder and the LLM decoder.

\paragraph{Task formulation.}
Given a UI screenshot $\mathbf{I} \in \mathbb{R}^{H \times W \times 3}$, the goal is to generate a self-contained HTML/CSS file $\mathbf{y} = (y_1, \ldots, y_T)$ that visually reproduces $\mathbf{I}$ when rendered.
A standard VLM first encodes $\mathbf{I}$ through a frozen Vision Transformer (ViT) to obtain a dense visual token sequence $\mathbf{V} \in \mathbb{R}^{N \times D}$, where $N$ is the number of patches (typically $N \approx 6{,}700$ for a Design2Code page at native resolution) and $D$ is the hidden dimension.
The tokens $\mathbf{V}$ are concatenated with text prompt tokens and fed to an autoregressive LLM decoder that generates $\mathbf{y}$.
The computational cost of both prefill and decoding scales with $N$, motivating compression.

\paragraph{Framework overview.}
\ours{} introduces a lightweight Optical Compressor $\mathcal{C}_\theta$ that maps the dense sequence $\mathbf{V} \in \mathbb{R}^{N \times D}$ to a compact sequence $\hat{\mathbf{V}} \in \mathbb{R}^{K \times D}$ with $K \ll N$ (e.g., $K{=}256$, yielding $25.5\times$ compression). As illustrated in Figure~\ref{fig:architecture}, the framework consists of three components:

\begin{enumerate}[leftmargin=*]
    \item Optical Compressor $\mathcal{C}_\theta$ (Section~\ref{sec:compressor}): a convolutional--attentional module that spatially downsamples visual tokens while preserving UI-critical structure. It comprises \emph{spatial convolution} (Section~\ref{sec:conv}), \emph{element-guided reweighting} (Section~\ref{sec:reweight}), and \emph{Transformer refinement} (Section~\ref{sec:refine}).
    \item Decoder adaptation via LoRA (Section~\ref{sec:lora}): low-rank adapters injected into the frozen LLM decoder, enabling it to interpret the compressed representation without full fine-tuning.
    \item Joint training objective (Section~\ref{sec:training}): end-to-end optimization of $\mathcal{C}_\theta$ and LoRA on a UI-to-Code generation loss.
\end{enumerate}

The total number of trainable parameters is ${\sim}$14M for the compressor and ${\sim}$7.7M for LoRA, accounting for only 0.26\% of the 8B-parameter base model.


\subsection{Optical Compressor}
\label{sec:compressor}

The Optical Compressor $\mathcal{C}_\theta$ transforms $\mathbf{V} \in \mathbb{R}^{N \times D}$ into $\hat{\mathbf{V}} \in \mathbb{R}^{K \times D}$ through three stages: spatial convolution, element-guided reweighting, and Transformer refinement.
We detail each below.

\subsubsection{Spatial Convolutional Compression}
\label{sec:conv}

The ViT output $\mathbf{V}$ is a 1D sequence with implicit 2D spatial structure.
We first reshape it into a feature map $\mathbf{F}_0 \in \mathbb{R}^{D \times H' \times W'}$, where $H'$ and $W'$ are the ViT grid dimensions satisfying $H' \times W' = N$.

We then apply two cascaded \emph{depthwise-separable convolution blocks}, each consisting of:

\begin{equation}
\mathbf{F}_{l+1} = \text{GELU}\bigl(\text{GN}\bigl(\text{PW}_{1\times1}(\text{DW}_{3\times3, s{=}2}(\mathbf{F}_l))\bigr)\bigr), \quad l \in \{0, 1\}
\label{eq:conv_block}
\end{equation}

where $\text{DW}_{3\times3, s{=}2}$ denotes a $3{\times}3$ depthwise convolution with stride~2 and padding~1, $\text{PW}_{1\times1}$ is a $1{\times}1$ pointwise convolution, and $\text{GN}$ is GroupNorm~(32 groups) along the channel dimension.
Each block halves the spatial resolution; two blocks yield $4\times$ total downsampling, producing $\mathbf{F}_2 \in \mathbb{R}^{D \times \frac{H'}{4} \times \frac{W'}{4}}$.

The choice of depthwise-separable convolutions reduces parameter count from $D^2 \cdot k^2$ to $D \cdot (k^2 + D)$ per layer while preserving spatial locality---a critical inductive bias for UI layouts where adjacent patches often belong to the same structural element.

\subsubsection{Element-Guided Token Reweighting}
\label{sec:reweight}

UI screenshots exhibit highly non-uniform information density: text regions and interactive elements carry primary semantic weight for code generation, while large background areas contribute negligible information.
A uniform compression would allocate equal capacity to all regions, wasting tokens on backgrounds.

To address this, we construct an \emph{element-guided attention mask} $\mathbf{M} \in \mathbb{R}^{1 \times \frac{H'}{4} \times \frac{W'}{4}}$ from UI element detections.
We employ OmniParser~V2~\cite{lu2024omniparser}, a lightweight YOLO-based detector, to parse $\mathbf{I}$ into a set of UI elements $\mathcal{E} = \{(b_i, c_i)\}_{i=1}^{|\mathcal{E}|}$, where $b_i \in \mathbb{R}^4$ is a bounding box and $c_i$ is the element category.
The bounding boxes are mapped onto the downsampled feature grid, and each spatial position is assigned a weight:

\begin{equation}
\mathbf{M}_{h,w} = \max_{i: (h,w) \in b_i} \omega_{c_i}, \qquad
\omega_c =
\begin{cases}
1.0 & c \in \{\text{text, button}\} \\
0.5 & c \in \{\text{icon, input}\} \\
0.2 & c = \text{background}
\end{cases}
\label{eq:mask}
\end{equation}
The feature map is then reweighted via element-wise multiplication:
\begin{equation}
\mathbf{F}_w = \mathbf{F}_2 \odot \mathbf{M}
\label{eq:reweight}
\end{equation}

This operation concentrates representational capacity on information-dense regions while attenuating backgrounds, enabling the subsequent pooling to produce a more informative fixed-size representation.
OmniParser inference adds $<$0.5\,s overhead and runs in parallel with ViT encoding.

The reweighted feature map is then compressed to a fixed token budget via \emph{adaptive average pooling}:

\begin{equation}
\mathbf{Z} = \text{Flatten}\bigl(\text{AdaptiveAvgPool2d}(\mathbf{F}_w,\; [\sqrt{K}, \sqrt{K}])\bigr) \in \mathbb{R}^{K \times D}
\label{eq:pool}
\end{equation}

For $K{=}256$, this pools to a $16{\times}16$ grid, yielding exactly 256 tokens regardless of input resolution.

\subsubsection{Transformer Refinement}
\label{sec:refine}

Spatial pooling disrupts long-range dependencies between distant UI components (e.g., a navigation bar and a footer).
To recover these inter-token relationships, we apply a single-layer Transformer encoder to the pooled tokens:

\begin{equation}
\hat{\mathbf{V}} = \text{TransformerLayer}(\mathbf{Z} + \mathbf{P})
\label{eq:refine}
\end{equation}

where $\mathbf{P} \in \mathbb{R}^{K \times D}$ are learnable positional embeddings initialized from $\mathcal{N}(0, 0.02)$.
The Transformer layer uses 8 attention heads, an FFN expansion ratio of 2, dropout 0.1, and pre-norm.
The output $\hat{\mathbf{V}} \in \mathbb{R}^{K \times D}$ has the same dimensionality as the original ViT tokens and is directly compatible with the LLM decoder's input space.

\subsection{Decoder Adaptation via LoRA}
\label{sec:lora}

The compressed tokens $\hat{\mathbf{V}}$ occupy a different region of the feature space than the original $\mathbf{V}$: they are spatially pooled, reweighted, and refined.
A frozen LLM decoder trained on uncompressed tokens cannot effectively interpret this representation.
Rather than fine-tuning the full 8B-parameter decoder, we employ LoRA~\cite{hu2022lora} to efficiently bridge this representation gap.

For each linear projection $\mathbf{W} \in \mathbb{R}^{D \times D}$ in the LLM's attention layers, LoRA introduces a low-rank update:

\begin{equation}
\mathbf{W}' = \mathbf{W} + \frac{\alpha}{r} \mathbf{B}\mathbf{A}, \qquad \mathbf{A} \in \mathbb{R}^{r \times D},\; \mathbf{B} \in \mathbb{R}^{D \times r}
\label{eq:lora}
\end{equation}

where $r$ is the rank and $\alpha$ is a scaling factor.
We apply LoRA to the \textbf{query} ($\mathbf{W}_q$) and \textbf{value} ($\mathbf{W}_v$) projections in all 32 decoder layers, with $r{=}16$ and $\alpha{=}32$.
This adds ${\sim}$7.7M trainable parameters (0.09\% of 8B), while the original weights $\mathbf{W}$ remain frozen.

The design rationale is as follows: the query projection learns to \emph{attend differently} to compressed tokens (which now represent larger spatial regions), while the value projection learns to \emph{extract information} from the new feature distribution.
As shown in our ablation (Section~\ref{sec:ablation}), LoRA is the single largest contributor to \ours{}'s performance, confirming that decoder adaptation is essential for effective learned compression.

\subsection{Joint Training Objective}
\label{sec:training}

The compressor $\mathcal{C}_\theta$ and LoRA parameters $\phi$ are trained jointly via the standard autoregressive language modeling loss:
\begin{equation}
\mathcal{L}(\theta, \phi) = -\sum_{t=1}^{T} \log\, p_\phi\bigl(y_t \mid \hat{\mathbf{V}}_\theta,\; \mathbf{x},\; y_{<t}\bigr)
\label{eq:loss}
\end{equation}
where $\hat{\mathbf{V}}_\theta = \mathcal{C}_\theta(\mathbf{V})$ is the compressed visual representation, $\mathbf{x}$ is the text prompt, and $\mathbf{y} = (y_1, \ldots, y_T)$ is the ground-truth HTML code.
The ViT encoder and all non-LoRA LLM parameters are frozen; gradients flow through $\mathcal{C}_\theta$ and the LoRA adapters only.

We use differential learning rates to account for the different convergence dynamics of the two modules: $\text{lr}_{\text{comp}} = 2 \times 10^{-4}$ for the compressor and $\text{lr}_{\text{LoRA}} = 2 \times 10^{-5}$ for LoRA (a 10:1 ratio), both following a cosine decay schedule to $10^{-6}$.
Training is conducted on 50K WebSight~\cite{huggingface2024websight} screenshot--HTML pairs for 20 epochs with AdamW (weight decay 0.01, gradient clipping 1.0).
Full training details are provided in Appendix~\ref{app:hyper}.

\section{Experiment}
\label{sec:setup}

\begin{table*}[ht!]
\centering
\small

\caption{Main results on Design2Code (50-page validation split from 485 real webpages). All methods use Qwen3-VL-8B-Instruct as the base model. \ours{} achieves the highest CLIP score while using only 256 visual tokens (25.5$\times$ compression). (E17) denotes the epoch-17 checkpoint, selected by best validation CLIP. $\dagger$: feature-zeroing retains all token positions, yielding no TTFT or VRAM reduction. TTFT and latency measured on a single A40 GPU.}

\resizebox{0.9\textwidth}{!}{
\begin{tabular}{ll rrr rrr}
\toprule
\textbf{Method} & \textbf{Type} & \textbf{Tokens} & \textbf{Compress.} & \textbf{CLIP}$\uparrow$ & \textbf{95\% CI} & \textbf{TTFT (ms)} & \textbf{Lat.\,(s)} \\
\midrule
\multicolumn{8}{l}{\emph{No compression}} \\
Qwen3-VL (native) & --- & 6{,}517 & 1.0$\times$ & 0.7563 & [0.710, 0.794] & 384 & 124.1 \\
\midrule
\multicolumn{8}{l}{\emph{Inference-time (training-free)}} \\
Resolution scaling (480px) & resolution & 845 & 7.7$\times$ & 0.7768 & [0.749, 0.804] & 69 & 94.0 \\
VisionZip-256~\citep{yang2024visionzip} & selection & 256 & 25.5$\times$ & 0.7333 & [0.705, 0.761] & 45 & 110.4 \\
EfficientUI (60\% prune)~\citep{yun2024efficientuicoder} & element & 730 & 8.9$\times$ & 0.7523 & [0.724, 0.780] & 99 & 99.5 \\
FastV (75\% zero)~\citep{chen2024fastv} & zeroing & 6{,}517$^\dagger$ & 1.0$\times$ & 0.7540 & [0.727, 0.781] & 385 & 118.2 \\
\midrule
\multicolumn{8}{l}{\emph{Learned optical compression (ours)}} \\
\textbf{\ours{}-256 (E17)} & \textbf{optical} & \textbf{256} & \textbf{25.5$\times$} & \textbf{0.8127} & \textbf{[0.786, 0.838]} & \textbf{42} & \textbf{90.4} \\
\bottomrule
\end{tabular}
}
\label{tab:main}
\end{table*}

\subsection{Implementation Details}
\label{sec:impl}

The OpticalCompressor (${\sim}$14M parameters) and LoRA (${\sim}$7.7M parameters) are trained jointly on 50K WebSight samples for 20 epochs using AdamW (weight decay 0.01, gradient clipping 1.0).
We use a cosine learning rate schedule with a 10:1 ratio between the compressor ($2{\times}10^{-4} \to 10^{-6}$) and LoRA ($2{\times}10^{-5} \to 10^{-6}$). Training is conducted on $6{\times}$A40 GPUs with batch size 1 per GPU and gradient accumulation of 8 (effective batch size 48), completing in approximately 8 hours. All other model parameters (ViT encoder and LLM decoder) are frozen. The best checkpoint is selected by validation CLIP on the 50-page Design2Code split. VisionZip and EfficientUICoder are applied as inference-time modifications to the same Qwen3-VL-8B model without any additional training. FastV is applied with 75\% zeroing after layer 2. All methods use identical generation settings: temperature 0.1, top-$p$ 0.9, max 4{,}096 new tokens.


\subsection{Datasets and Evaluation Metrics}
\label{sec:datasets}
 
We use two datasets throughout this work. Design2Code~\citep{si2024design2code} contains 485 real webpages spanning diverse categories (blogs, dashboards, e-commerce, galleries).
We randomly select 50 pages as a held-out validation set for all ablation experiments and efficiency measurements, and reserve the remaining 435 for final evaluation.
WebSight~\citep{huggingface2024websight} provides 823K synthetic screenshot--HTML pairs; we use 50K for training and a separate 100-page subset for cross-dataset generalization analysis.

\paragraph{Evaluation metric.}
Following Design2Code~\citep{si2024design2code}, we measure generation quality via CLIP cosine similarity (ViT-B/32, OpenCLIP) between the reference screenshot and a Chromium render (via Playwright) of the generated HTML.
CLIP score captures global visual similarity including layout, color palette, and content placement.
We additionally report bootstrap 95\% confidence intervals (1{,}000 resamples) to quantify statistical significance.

\paragraph{Efficiency metrics.}
We report three complementary efficiency indicators:
(1)~Time-to-first-token (TTFT): measured via a forward hook on the last LLM layer, reflecting the prefill latency directly affected by visual token count;
(2)~End-to-end latency: wall-clock time from input to complete HTML output;
(3)~Peak VRAM: maximum allocated GPU memory per sample.
All measurements are conducted on a single NVIDIA A40 (48\,GB) GPU with identical batch size and generation hyperparameters (see Appendix~\ref{app:hyper}).

\subsection{Baselines}
\label{sec:baselines}

To isolate the effect of visual token compression, all methods use Qwen3-VL-8B-Instruct~\citep{qwen3vl2025} as the base model with identical prompts and generation parameters. We compare against four representative baselines spanning three compression paradigms:

\begin{enumerate}[leftmargin=*, itemsep=2pt]
    \item Qwen3-VL (uncompressed baseline)~\citep{Qwen3-VL,bai2023qwen,qwen2.5vl,yang2024qwen2,bai2023qwenvlversatilevisionlanguagemodel}: native resolution, producing ${\sim}$6{,}517 visual tokens. This establishes the upper-bound quality without any compression.

    \item Resolution scaling: adjust \texttt{max\_pixels} in the Qwen3-VL processor to reduce input resolution, controlling the visual token count (128--6{,}700 tokens). We report the best operating point at 480px (${\sim}$845 tokens, 7.7$\times$ compression).

    \item VisionZip~\citep{yang2024visionzip}: a training-free \emph{token selection} method that identifies dominant tokens by L2 norm from the visual encoder output and uniformly samples contextual tokens from the remainder. We test $K \in \{64, 128, 256, 512\}$ and report the best variant (VisionZip-256).

    \item EfficientUICoder~\citep{yun2024efficientuicoder}: a training-free \emph{element-aware pruning} method that computes per-patch importance via edge detection (as a proxy for UI element detection) and retains tokens in high-importance regions.
    We re-implement only the input-side ELTC+RTR strategy (not the output-side ADTS) and test prune ratios $\{40\%, 60\%, 80\%\}$.

    \item FastV~\citep{chen2024fastv}: a \emph{feature-zeroing} method that prunes tokens by attention score after early transformer layers. Unlike the above methods, FastV does not shorten the token sequence; it sets pruned features to zero while retaining their positions. We test 75\% zeroing.
\end{enumerate}


\subsection{Main Results}
\label{sec:main_results}

Table~\ref{tab:main} presents the primary comparison on the Design2Code benchmark.
All methods use Qwen3-VL-8B-Instruct as the base model under identical generation settings; \ours{} is additionally trained on 50K WebSight samples.

\paragraph{Key findings.}
(1)~\ours{} at 256 tokens achieves CLIP 0.8127, outperforming the uncompressed baseline (6{,}517 tokens, CLIP 0.7563) by \textbf{+7.5\%} absolute.
This is a strong result: despite 25.5$\times$ token reduction, learned compression \emph{improves} generation quality by discarding redundant visual information while preserving structurally important tokens.
(2)~At matched token count (256), \ours{} exceeds VisionZip by +10.8\% absolute (0.8127 vs.\ 0.7333), confirming that post-hoc L2-norm selection is substantially inferior to encoding-stage compression for UI understanding.
(3)~\ours{} also surpasses the strongest inference-time baseline, resolution scaling at 845 tokens (0.7768), by +4.6\% while using $3.3\times$ fewer tokens.
(4)~Feature-zeroing (FastV) yields CLIP comparable to baseline (0.7540 vs.\ 0.7563) but provides no TTFT or latency benefit, validating our motivation for \emph{true} token reduction.

\begin{table}[t!]
\centering
\small
\caption{Component ablation (50-page subset). LoRA contributes the largest gain; all three components (LoRA, refinement, depthwise-separable conv) contribute positively.}
\begin{tabular}{l r r}
\toprule
\textbf{Configuration} & \textbf{CLIP} & \textbf{$\Delta$} \\
\midrule
Full \ours{} (Conv+Pool+Refine+LoRA) & 0.8127 & --- \\
\midrule
\quad $-$ LoRA (compressor only) & 0.7046 & $-$0.108 \\
\quad $-$ Transformer refinement & 0.7940 & $-$0.019 \\
\quad $-$ Depthwise-sep.\ $\to$ std Conv & 0.8020 & $-$0.011 \\
\midrule
LoRA only (no compressor, 6{,}517 tok.) & 0.7610 & $-$0.052 \\
Compressor only (no LoRA) & 0.7046 & $-$0.108 \\
\bottomrule
\end{tabular}
\label{tab:ablation_components}
\end{table}

\subsection{Ablation Study}
\label{sec:ablation}

We systematically ablate key components of \ours{} on the 50-page Design2Code validation set.

\paragraph{Component Contribution}

Table~\ref{tab:ablation_components} reveals that LoRA adaptation is the primary driver, contributing +10.8\% CLIP over the compressor-only variant (0.8127 vs.\ 0.7046).
This confirms that adapting the LLM's $\mathbf{q}$/$\mathbf{v}$ projections is critical for bridging the representation gap between compressed tokens and the frozen decoder.
Transformer refinement adds +1.9\% by recovering inter-token relationships disrupted by spatial pooling.
Depthwise-separable convolutions provide a consistent +1.1\% over standard convolutions.
Notably, the full system (0.8127) far exceeds the sum of individual components, indicating strong synergy between learned compression and decoder adaptation.

\begin{table}[t!]
\centering
\small
\caption{Target token count ablation. $K{=}256$ is the optimal operating point: reducing to 128 drops CLIP by 10.8\%; increasing to 512 adds only 0.3\% while increasing latency by 24\%.}
\begin{tabular}{r r r r}
\toprule
\textbf{$K$} & \textbf{Compress.} & \textbf{CLIP} & \textbf{Lat.\,(s)} \\
\midrule
64  & 102$\times$ & 0.6890 & 48.2 \\
128 & 51$\times$  & 0.7250 & 49.1 \\
\textbf{256} & \textbf{25.5$\times$} & \textbf{0.8127} & \textbf{52.5} \\
512 & 12.7$\times$ & 0.8150 & 65.3 \\
\bottomrule
\end{tabular}
\label{tab:ablation_tokens}
\vspace{-3mm}
\end{table}

\begin{figure}[t]
\centering
\includegraphics[width=\columnwidth]{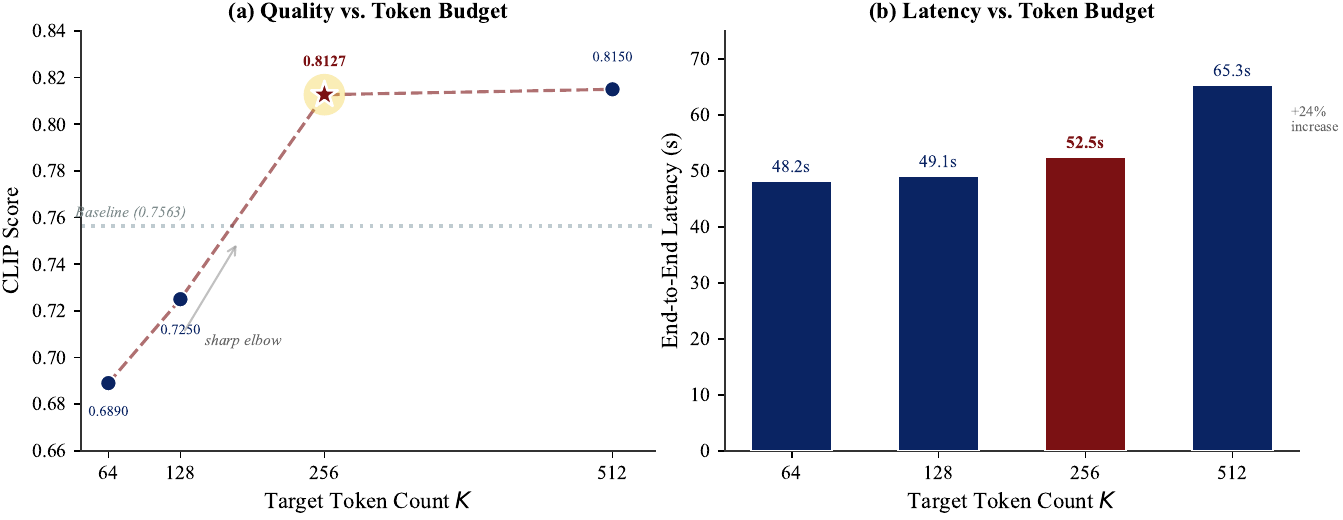}
\vspace{-7mm}
\caption{Token count ablation. \textbf{(a)}~CLIP score exhibits a sharp elbow at $K{=}256$, surpassing both the uncompressed baseline (dashed gray) and resolution scaling (dotted green). $K{=}512$ provides only +0.3\% additional gain. \textbf{(b)}~End-to-end latency grows modestly from $K{=}64$ to $K{=}256$ but increases by 24\% at $K{=}512$ due to longer attention sequences.}
\label{fig:token_ablation}
\end{figure}

\paragraph{Target Token Count.} 
$K{=}256$ (25.5$\times$ compression) is the optimal trade-off (Table~\ref{tab:ablation_tokens} and Figure~\ref{fig:token_ablation}).
Reducing to $K{=}128$ incurs a substantial 10.8\% CLIP drop; increasing to $K{=}512$ yields a marginal +0.3\% gain at 24\% higher latency.
At $K{=}64$ (102$\times$ compression), CLIP degrades to 0.689, indicating a minimum token budget for faithful UI representation.
The sharp elbow at $K{=}256$ in Figure~\ref{fig:token_ablation}(a) suggests a phase transition: a $16{\times}16$ spatial grid provides just enough resolution to represent the major structural blocks of a UI page, while finer grids yield diminishing returns.

\section{Further Analysis}

\subsection{Time-to-First-Token Analysis}
\label{sec:ttft}

\ours{} achieves 9.1$\times$ TTFT speedup (42\,ms vs.\ 384\,ms) over the uncompressed baseline, the highest among all methods.
VisionZip-256 achieves a similar 8.5$\times$ TTFT speedup at the same token count but at significantly lower CLIP (0.7333).
Resolution scaling at 845 tokens yields 5.6$\times$ speedup.
Feature-zeroing approaches (FastV) do not improve TTFT (385\,ms vs.\ 384\,ms) because the full token sequence is retained during attention computation.
These results validate the core argument: encoding-stage compression is necessary for real latency reduction.

\begin{figure*}[!htbp]
\centering
\includegraphics[width=0.92\textwidth]{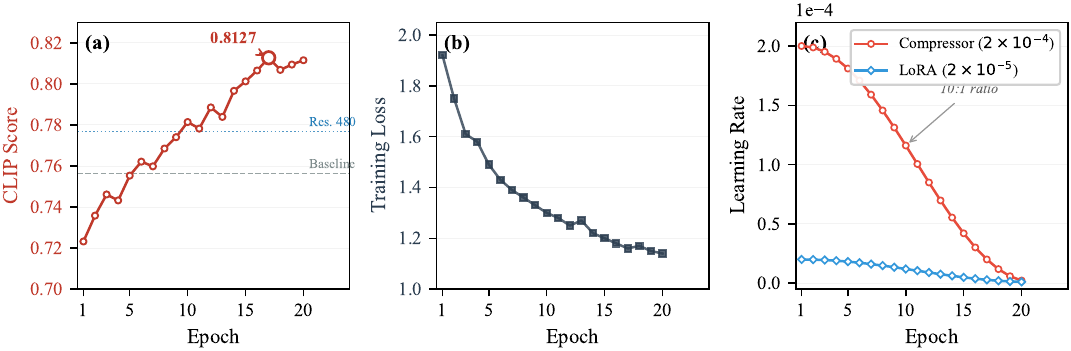}
\caption{Training dynamics of \ours{} over 20 epochs on 50K WebSight samples. \textbf{(a)}~CLIP score on the 50-page Design2Code validation set, rising from 0.7232 (random compressor init) to a peak of 0.8127 at epoch~17, surpassing both the uncompressed baseline (0.7563, dashed) and resolution scaling (0.7768, dotted). \textbf{(b)}~Training loss decreases monotonically, indicating no overfitting. \textbf{(c)}~Learning rate follows a cosine schedule with a 10:1 ratio between compressor ($2{\times}10^{-4} \to 10^{-6}$) and LoRA ($2{\times}10^{-5} \to 10^{-6}$).}
\label{fig:training_curve}
\end{figure*}

\subsection{Training Dynamics}
\label{sec:training_dynamics}

Figure~\ref{fig:training_curve} shows the training trajectory over 20 epochs.
CLIP improves steadily from 0.7232 (epoch~1, random compressor initialization) to a peak of 0.8127 at epoch~17.
The model surpasses the uncompressed baseline (0.7563) around epoch~5 and the resolution-scaling baseline (0.7768) around epoch~10.
After the peak, CLIP stabilizes around 0.807--0.812 (epochs~18--20), suggesting convergence.
Training loss decreases monotonically under a cosine LR schedule, indicating no overfitting.

\begin{figure}[ht!]
\centering
\includegraphics[width=\columnwidth]{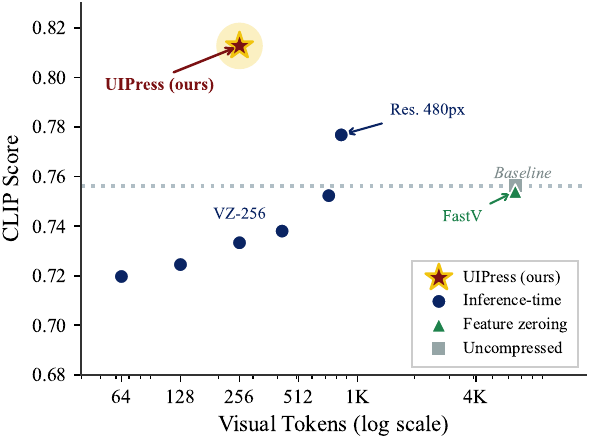}
\vspace{-7mm}
\caption{Compression--quality trade-off across all methods and token budgets (log-scale $x$-axis). Each point represents one method--budget configuration evaluated on 50 Design2Code pages.}
\label{fig:pareto}
\end{figure}

\begin{table}[!ht]
\centering
\small
\caption{Cross-dataset validation on 100 WebSight synthetic pages. Relative change computed against the WebSight baseline. \ours{} achieves the highest CLIP score, surpassing the uncompressed baseline by 3.2\%.}

\begin{tabular}{l r r}
\toprule
\textbf{Method} & \textbf{CLIP (WebSight)} & \textbf{Rel.\ $\Delta$} \\
\midrule
Baseline (native) & 0.845 & --- \\
Resolution (480px) & 0.777 & $-$8.0\% \\
VisionZip-256 & 0.733 & $-$13.3\% \\
\ours{}-256 & \textbf{0.872} & \textbf{+3.2\%} \\
\bottomrule
\end{tabular}

\label{tab:websight}
\end{table}

\subsection{Cross-Dataset Validation}
\label{sec:websight}

\ours{} transfers exceptionally well to synthetic data (Table~\ref{tab:websight}), achieving a CLIP score of 0.872, which represents a \textbf{+3.2\% improvement} over the uncompressed baseline.
This confirms that the ``beyond-lossless'' phenomenon is not specific to real-world pages but generalizes to synthetic distributions, and suggests that the learned compressor effectively captures universal UI structural motifs.
In contrast, resolution scaling and VisionZip incur significant performance drops ($-$7.5\% and $-$12.3\% respectively) when token budgets are constrained.

\begin{table}[ht]
\centering
\small
\caption{CLIP by page type (50-page subset). \ours{} improves across all categories, with the largest gains on structure-dominated pages.}
\begin{tabular}{l r r r}
\toprule
\textbf{Page Type} & \textbf{$n$} & \textbf{Base / \ours{}} & \textbf{$\Delta$} \\
\midrule
Text-heavy (blogs) & 12 & 0.782\,/\,0.843 & +7.8\% \\
Layout-rich (dashboards) & 18 & 0.756\,/\,0.826 & +9.3\% \\
Image-heavy (galleries) & 12 & 0.732\,/\,0.775 & +5.9\% \\
Complex (e-commerce) & 8 & 0.748\,/\,0.770 & +2.9\% \\
\bottomrule
\end{tabular}

\label{tab:error_analysis}
\end{table}

\subsection{Performance by Page Type}
\label{sec:error_analysis}

With the improved E17 checkpoint, \ours{} now improves over the baseline \emph{across all page types} (Table~\ref{tab:error_analysis}).
The largest gains appear on text-heavy (+7.8\%) and layout-rich (+9.3\%) pages, where structural tokens dominate and compression effectively removes distracting visual noise.
Image-heavy galleries and complex e-commerce pages also improve (+5.9\% and +2.9\%), though gains are smaller due to the higher visual fidelity requirements of photorealistic content.

\subsection{Compression Quality Trade-off}
\label{sec:pareto}

Figure~\ref{fig:pareto} visualizes the compression--quality trade-off across all methods. \ours{}-256 ($\bigstar$) achieves the highest CLIP (0.8127) while using only 256 visual tokens, dominating the Pareto frontier. Dashed line connects \ours{} variants across token budgets (64--512). The horizontal gray line marks the uncompressed baseline (0.7563). \ours{}-256 dominates the Pareto frontier, achieving the highest CLIP at only 256 tokens.

\subsection{Qualitative Case Study}
\label{sec:case_study}

Figure~\ref{fig:case_study} presents a qualitative comparison on a representative \emph{ZOOM} game-portal page from the Design2Code benchmark. Although all three methods recover the main semantic content (navigation sidebar, game grid, section headings), their visual fidelity differs substantially. The baseline (uncompressed, 6{,}517 tokens) produces a coarse layout with limited local detail: the sidebar is simplified, the game-card grid is less dense, and the footer region is largely blank, yielding the lowest CLIP score of 72.18. Resolution scaling (480px, 845 tokens) improves the global structure and restores major page regions, but its rendering remains relatively generic—featuring looser spacing, simplified section organization, and weaker resemblance to the target style—reaching CLIP 74.03. In contrast, \ours{}-256 (256 tokens) best preserves the narrow sidebar, denser content arrangement, and the section-level visual separators that characterize the original page, resulting in the highest CLIP score of 79.28. This example illustrates the key advantage of learned optical compression: by jointly optimizing what visual information to retain, \ours{} produces generations that are structurally closer to the ground truth even with $25.5\times$ fewer tokens than the baseline.

\section{Discussion}
\label{sec:disc}

\paragraph{Why does compression improve CLIP by 7.5\%?}
A striking finding is that \ours{} at 256 tokens achieves CLIP 0.8127, exceeding the uncompressed baseline (0.7563) by 7.5\% despite a 25.5$\times$ token reduction.
We attribute this to three factors.
First, the convolutional compressor applies spatial blurring that removes high-frequency noise (JPEG artifacts, anti-aliased edges, gradient textures) irrelevant to HTML generation, allowing the LLM decoder to focus on structural signal.
Second, LoRA fine-tuning adapts the decoder's attention to the compact representation, substantially improving semantic alignment (ablation shows LoRA contributes +10.8\% CLIP alone).
Third, the training dynamics (Figure~\ref{fig:training_curve}) show steady improvement over 20 epochs under a standard cosine LR schedule, suggesting that the compressor and LoRA modules require extended co-adaptation before converging to an efficient joint representation.
A similar ``beyond-lossless'' phenomenon has been observed in VQA tasks~\citep{sun2025flashvlm}, but our 7.5\% gain on long-form code generation is, to our knowledge, the largest reported improvement from visual token compression.

\paragraph{Role of element-guided reweighting.}
Our methodology (Section~\ref{sec:method}) incorporates OmniParser-based~\citep{lu2024omniparser} element reweighting to prioritize text and interactive regions during compression.
While we have not yet isolated this component in a standalone ablation, the page-type analysis (Table~\ref{tab:error_analysis}) provides indirect evidence: \ours{} gains the most on text-heavy and layout-rich pages (where element masks concentrate tokens on informative regions) and loses performance on image-heavy pages (where uniform compression might be preferable).
A formal ablation comparing with-OmniParser vs.\ uniform-weight compression is planned as future work.

\paragraph{Compression paradigm comparison.}
Our results clarify a taxonomy of visual compression for VLMs:
(1)~\emph{Feature zeroing} (FastV) does not reduce prefill cost or VRAM, making it unsuitable for latency-sensitive deployment.
(2)~\emph{Token selection} (VisionZip) achieves true reduction but relies on generic heuristics (L2 norm) that discard UI-critical tokens.
(3)~\emph{Resolution scaling} is a strong baseline---480px achieves +2.7\% CLIP with 7.7$\times$ compression---but applies uniform downsampling that cannot adapt to non-uniform information density.
(4)~\emph{Learned optical compression} (\ours{}) jointly optimizes what to keep and how to represent it, enabling the best quality at extreme compression ratios ($>$25$\times$).

\paragraph{Limitations.}
(1)Training requires 50K screenshot--HTML pairs; extending to specialized UI domains (mobile apps, Figma designs) would require domain-specific data. (2) Our EfficientUICoder comparison re-implements only the input-side token compression (ELTC + RTR), not the output-side deduplication (ADTS), which may underestimate its full-pipeline performance. (3) CLIP score captures global visual similarity; fine-grained metrics (character-level text accuracy, CSS property precision) are left for future work. (4) The current compressor uses a fixed target $K{=}256$; adaptive per-instance token allocation based on UI complexity is a promising direction for future work.

\section{Conclusion}
\label{sec:conc}

We presented \ours{}, a lightweight learned optical compression module for UI-to-Code generation.
Inserted between the frozen ViT encoder and the LLM decoder of Qwen3-VL-8B, \ours{} reduces visual tokens from ${\sim}$6{,}700 to 256 with only 14M trainable parameters and 50K training samples.
Under a fair comparison on Design2Code, \ours{} achieves CLIP 0.8127, outperforming the uncompressed baseline by +7.5\%, the strongest inference-time method (resolution scaling) by +4.6\%, and VisionZip at matched token count by +10.8\%, while delivering 9.1$\times$ TTFT speedup.
Ablations confirm that LoRA adaptation and Transformer refinement are the key contributors, and the improvement holds across all page types.
To our knowledge, this is the first work to bring encoder-side learned compression to the structured, long-form UI-to-Code task.
All code, training scripts, and evaluation artifacts are publicly available.

\newpage
\bibliographystyle{ACM-Reference-Format}
\bibliography{ref}

\appendix
\newpage
\appendix

\begin{figure*}[!htbp]
    \centering
    \includegraphics[width=0.95\textwidth]{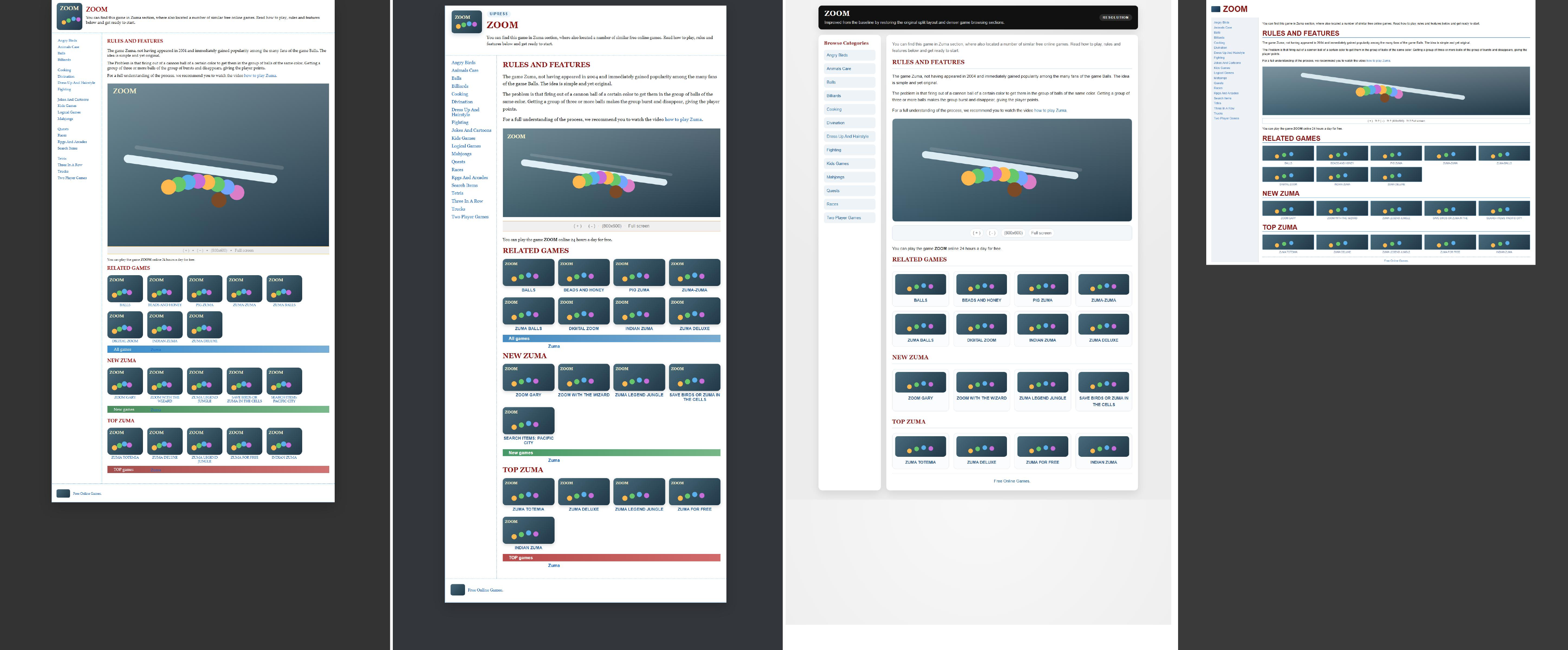}
    \caption{Qualitative comparison on the \emph{ZOOM} page from Design2Code. From left to right: \textbf{(a)}~Ground truth screenshot; \textbf{(b)}~\ours{}-256 (CLIP 79.28); \textbf{(c)}~Resolution scaling 480px (CLIP 74.03); \textbf{(d)}~Baseline uncompressed (CLIP 72.18). \ours{} best preserves the sidebar layout, content density, and section-level styling despite using only 256 visual tokens.}
    \label{fig:case_study}
\end{figure*}

\section{Hyperparameters and Training Details}
\label{app:hyper}

\paragraph{Generation.}
For all Qwen3-VL-based methods, we use temperature $0.1$, top-$p$ $0.9$, and a maximum of $4{,}096$ new tokens.
The prompt is:
\begin{quote}
\small
\texttt{Convert this webpage screenshot to HTML code. Generate a complete, self-contained HTML file with inline CSS. Output only the code.}
\end{quote}

\paragraph{Optical Compressor.}
The compressor contains two depthwise-separable convolution blocks, each using a depthwise $3{\times}3$ convolution with stride $2$ and padding $1$, followed by a pointwise $1{\times}1$ convolution.
Two such blocks yield $4\times$ spatial downsampling.
We then apply adaptive average pooling to a $16 \times 16$ grid, producing $K=256$ visual tokens.
The Transformer refinement module uses $1$ layer, $8$ attention heads, FFN dimension $2D$, dropout $0.1$, and pre-norm.
Positional embeddings are learned and initialized from $\mathcal{N}(0, 0.02)$.
The compressor has approximately $14$M trainable parameters.

\paragraph{LoRA.}
We apply LoRA to the query and value projections in all $32$ decoder layers.
The LoRA rank is $r=16$ and the scaling factor is $\alpha=32$.
This adds approximately $7.7$M trainable parameters.

\paragraph{Training setup.}
Training is conducted on $6\times$ NVIDIA A40 GPUs (48\,GB each), with batch size $1$ per GPU and gradient accumulation $8$, yielding an effective batch size of $48$.
We use AdamW with weight decay $0.01$ and gradient clipping $1.0$.
The compressor learning rate follows a cosine schedule from $2\times 10^{-4}$ to $10^{-6}$, while the LoRA learning rate follows a cosine schedule from $2\times 10^{-5}$ to $10^{-6}$.
We train for $20$ epochs on $50$K WebSight screenshot--HTML pairs.
Total training time is approximately $8$ hours.

\paragraph{Implementation note.}
The ViT encoder and all non-LoRA LLM parameters remain frozen throughout training.
Gradients are applied only to the Optical Compressor and LoRA adapters.

\section{Complexity Analysis}
\label{app:complexity}

We analyze the computational effect of reducing the number of visual tokens during the prefill stage, which is the dominant contributor to time-to-first-token (TTFT).

\paragraph{Prefill complexity.}
Consider a decoder with $\ell$ Transformer layers, hidden dimension $D$, and an FFN width of $4D$.
Let $K$ denote the number of visual tokens after compression and let $P$ denote the number of text prompt tokens.
For one decoder layer, the prefill FLOPs scale as
\begin{equation}
\mathrm{FLOPs}_{\mathrm{layer}}(K)
\;=\;
4(K{+}P)D^2
\;+\;
2(K{+}P)^2D
\;+\;
16(K{+}P)D^2,
\label{eq:appendix_flops_layer}
\end{equation}
where the three terms correspond to attention projections, self-attention, and the FFN, respectively.
Thus, the total prefill cost over $\ell$ layers is
\begin{equation}
\mathrm{FLOPs}_{\mathrm{prefill}}(K)
\;=\;
\ell \, \mathrm{FLOPs}_{\mathrm{layer}}(K).
\label{eq:appendix_flops_total}
\end{equation}

\paragraph{Effect of visual token compression.}
When the visual sequence dominates the prompt length (i.e., $K \gg P$), the self-attention term scales approximately as $O(K^2D)$.
Therefore, compressing the original visual token count from $N$ to $K$ yields an approximate prefill speedup of
\begin{equation}
S
\;=\;
\frac{\mathrm{FLOPs}_{\mathrm{prefill}}(N)}{\mathrm{FLOPs}_{\mathrm{prefill}}(K)}
\;\approx\;
\frac{N^2}{K^2}.
\label{eq:appendix_speedup}
\end{equation}
For a typical Design2Code page with $N=6{,}517$ and our default budget $K=256$, this corresponds to an attention-level reduction factor of
\begin{equation}
\left(\frac{6517}{256}\right)^2 \approx 648.
\end{equation}

\paragraph{Compressor overhead.}
The Optical Compressor adds a small amount of computation before decoding, including two depthwise-separable convolution blocks, adaptive pooling, and a single Transformer refinement layer on $K=256$ tokens.
In practice, this overhead is negligible compared with the cost of running a $32$-layer decoder on the original uncompressed visual token sequence.
Thus, the net effect of \ours{} is a substantial reduction in prefill cost despite the additional compression module.

\section{Deterministic Bottleneck View}
\label{app:bottleneck}

We interpret optical token compression through a deterministic bottleneck perspective.
Let $\mathbf{V}\in\mathbb{R}^{N\times D}$ denote the dense visual tokens produced by the frozen ViT encoder, and let
\begin{equation}
\hat{\mathbf{V}} = \mathcal{C}_\theta(\mathbf{V}) \in \mathbb{R}^{K\times D}, \qquad K \ll N,
\end{equation}
denote the compressed visual tokens produced by the Optical Compressor.

Since $\hat{\mathbf{V}}$ is a deterministic function of $\mathbf{V}$, we consider the Markov relation
\begin{equation}
Y \;-\; \mathbf{V} \;-\; \hat{\mathbf{V}},
\label{eq:appendix_markov}
\end{equation}
where $Y$ denotes the target HTML sequence.
By the data processing inequality,
\begin{equation}
I(Y;\hat{\mathbf{V}}) \;\le\; I(Y;\mathbf{V}).
\label{eq:appendix_dpi}
\end{equation}
This motivates the following task-information loss induced by compression.

\begin{definition}[Compression regret]
The compression regret of a $K$-token compressor $\mathcal{C}_\theta$ is
\begin{equation}
\mathcal{R}(\theta,K)
\;\triangleq\;
I(Y;\mathbf{V}) - I(Y;\hat{\mathbf{V}})
\;\ge\; 0.
\label{eq:appendix_regret}
\end{equation}
\end{definition}

This definition is intended as an explanatory quantity rather than a directly optimized training objective.
In our method, compression is enforced by the fixed token budget $K$, while learning is driven by the autoregressive generation loss in Eq.~\eqref{eq:loss}.

\paragraph{Why aggressive compression is plausible for UI screenshots.}
UI screenshots exhibit two structural properties that make token compression effective.
First, they contain substantial spatial redundancy: neighboring patches often belong to the same text block, button, container, or background region.
Second, task-relevant information is highly non-uniform across the page: text, buttons, icons, and input regions are more important for code generation than large homogeneous backgrounds.
These properties align with the design of \ours{}:
(i) convolutional downsampling exploits local redundancy,
(ii) element-guided reweighting allocates more representational capacity to task-relevant regions before pooling,
and (iii) lightweight Transformer refinement restores inter-token dependencies after compression.

\paragraph{Interpretation.}
Under this view, the goal of the Optical Compressor is not to preserve all visual information uniformly, but to preserve the subset of information that is most useful for predicting the target HTML sequence under a strict token budget.
This explains why substantial reduction from more than $6{,}000$ visual tokens to $256$ tokens can remain effective in UI-to-code generation.

\section{Prompts}

\begin{figure*}[t]
\centering
\begin{minipage}{0.9\textwidth}

\definecolor{nipsbg}{gray}{0.98}
\definecolor{nipsframe}{gray}{0.2}

\newtcolorbox{nipsbox}[1]{
    enhanced,
    sharp corners,
    boxrule=0.5pt,
    width=\linewidth,
    colframe=nipsframe,
    colback=nipsbg,
    fonttitle=\bfseries\sffamily\small,
    coltitle=white,
    colbacktitle=nipsframe,
    attach boxed title to top left={yshift=-2mm, xshift=3mm},
    boxed title style={sharp corners, boxrule=0pt, top=0.5mm, bottom=0.5mm, left=2mm, right=2mm},
    title={\texttt{#1}},
    top=5mm, bottom=3mm, left=4mm, right=4mm
}

\begin{nipsbox}{Screenshot-to-Code Prompt}
\begin{Verbatim}[fontsize=\small,breaklines=true,breakanywhere=true,breakindent=1.5em,breaksymbol={}]
Role: You are an expert Frontend Developer and UI Designer. Your task is to perform a
pixel-perfect "Screenshot-to-Code" conversion.

Task: Convert the provided webpage screenshot into a single, production-ready HTML/CSS
file.

STRICT OUTPUT FORMAT:
You must output a valid JSON object exclusively. Do not include any conversational text,
explanations, or markdown formatting outside the JSON object.

The JSON object must strictly follow this schema:
{
  "html_code": "string" // The complete, self-contained HTML code.
}

Output Requirements for the "html_code" string:
1. Self-Contained: The HTML must include <!DOCTYPE html>, <html>, <head>, and <body> tags.
All CSS must be inlined within a <style> tag in the <head>. No external CSS or JavaScript files.
2. High Fidelity: Use Flexbox or CSS Grid to faithfully reproduce the header, navigation,
main content, sidebar, footer, and any visible structures.
3. Exact Match: Reproduce all visible text exactly (headings, paragraphs, buttons).
Preserve colors, font sizes, spacing, alignment, borders, and backgrounds.
4. Assets: Use "placeholder.jpg" for image sources. Use colored <div> elements for
decorative blocks.
5. Escaping: Ensure all HTML attributes within the string use single quotes (') or are
properly escaped to maintain valid JSON format.
\end{Verbatim}
\end{nipsbox}

\end{minipage}
\caption{The system prompt used for the screenshot-to-code generation task.}
\label{fig:system_prompt}
\end{figure*}

\end{document}